\documentclass[11pt,a4paper]{article}
\usepackage[hyperref]{acl2019}
\usepackage{times}
\usepackage{latexsym}
\usepackage{url}

\usepackage{balance}

\usepackage{array}
\usepackage{amssymb}
\usepackage{amsmath}
\usepackage{bm}
\usepackage{color}
\usepackage{multirow}
\usepackage{subfig}
\usepackage{arydshln}
\usepackage{amsfonts}
\usepackage{graphicx} 
\usepackage{graphics}
\usepackage{CJKutf8}
\usepackage{standalone}

\usepackage{caption}
\aclfinalcopy 

\title{Exploiting Sentential Context for Neural Machine Translation}

\author{
Xing Wang \\ Tencent AI Lab \\ {\tt brightxwang@tencent.com}   \And
Zhaopeng Tu \\ Tencent AI Lab \\ {\tt zptu@tencent.com} \AND
Longyue Wang \\ Tencent AI Lab \\ {\tt vinnylywang@tencent.com} \And
Shuming Shi \\ Tencent AI Lab \\ {\tt shumingshi@tencent.com}
}

\begin{document}
\maketitle
\begin{abstract}
In this work, we present novel approaches to exploit sentential context for neural machine translation (NMT). Specifically,  we first show that a {\em shallow sentential context} extracted from the top encoder layer only, can improve translation performance via contextualizing the encoding representations of individual words. Next, we introduce a {\em deep sentential context}, which aggregates the sentential context representations from all the internal layers of the encoder to form a more comprehensive context representation. Experimental results on the WMT14 English$\Rightarrow$German and English$\Rightarrow$French benchmarks show that our model consistently improves performance over the strong \textsc{Transformer} model~\cite{Vaswani:2017:NIPS}, demonstrating the necessity and effectiveness of exploiting sentential context for NMT.
\end{abstract}

\section{Introduction}

Sentential context, which involves deep syntactic and semantic structure of the source and target languages~\cite{Nida:1969}, is crucial for machine translation. In statistical machine translation (SMT), the sentential context has proven beneficial for predicting local translations~\cite{meng-EtAl:2015:ACL-IJCNLP, zhang2015local}. The exploitation of sentential context in neural machine translation ~\citep[NMT,][]{bahdanau2015neural}, however, is not well studied. Recently,~\newcite{C18-1276} showed that the translation at each time step should be conditioned on the whole {\em target-side} context. They introduced a deconvolution-based decoder to provide the global information from the target-side context for guidance of decoding.

In this work, we propose simple yet effective approaches to exploiting {\em source-side} global sentence-level context for NMT models. 
We use encoder representations to represent the source-side context, which are summarized into a {\em sentential context} vector. The source-side context vector is fed to the decoder, so that translation at each step is conditioned on the whole {\em source-side} context.
Specifically, we propose two types of sentential context: 1) the {\em shallow} one that only exploits the top encoder layer, and 2) the {\em deep} one that aggregates the sentence representations of all the encoder layers. The deep sentential context can be viewed as a more comprehensive global sentence representation, since different types of syntax and semantic information are encoded in different encoder layers \cite{D16-1159,Peters:2018:NAACL,W18-5431}.

We validate our approaches on top of the state-of-the-art \textsc{Transformer} model~\cite{Vaswani:2017:NIPS}. Experimental results on the benchmarks WMT14 English$\Rightarrow$German and English$\Rightarrow$French translation tasks show that exploiting sentential context consistently improves translation performance across language pairs.
Among the model variations, the deep strategies consistently outperform their shallow counterparts, which confirms our claim.
Linguistic analyses~\cite{P18-1198} on the learned representations reveal that the proposed approach indeed provides richer linguistic information.

\begin{figure*}[t]
\centering
\subfloat[Vanilla]{
\includegraphics[height=0.20\textwidth]{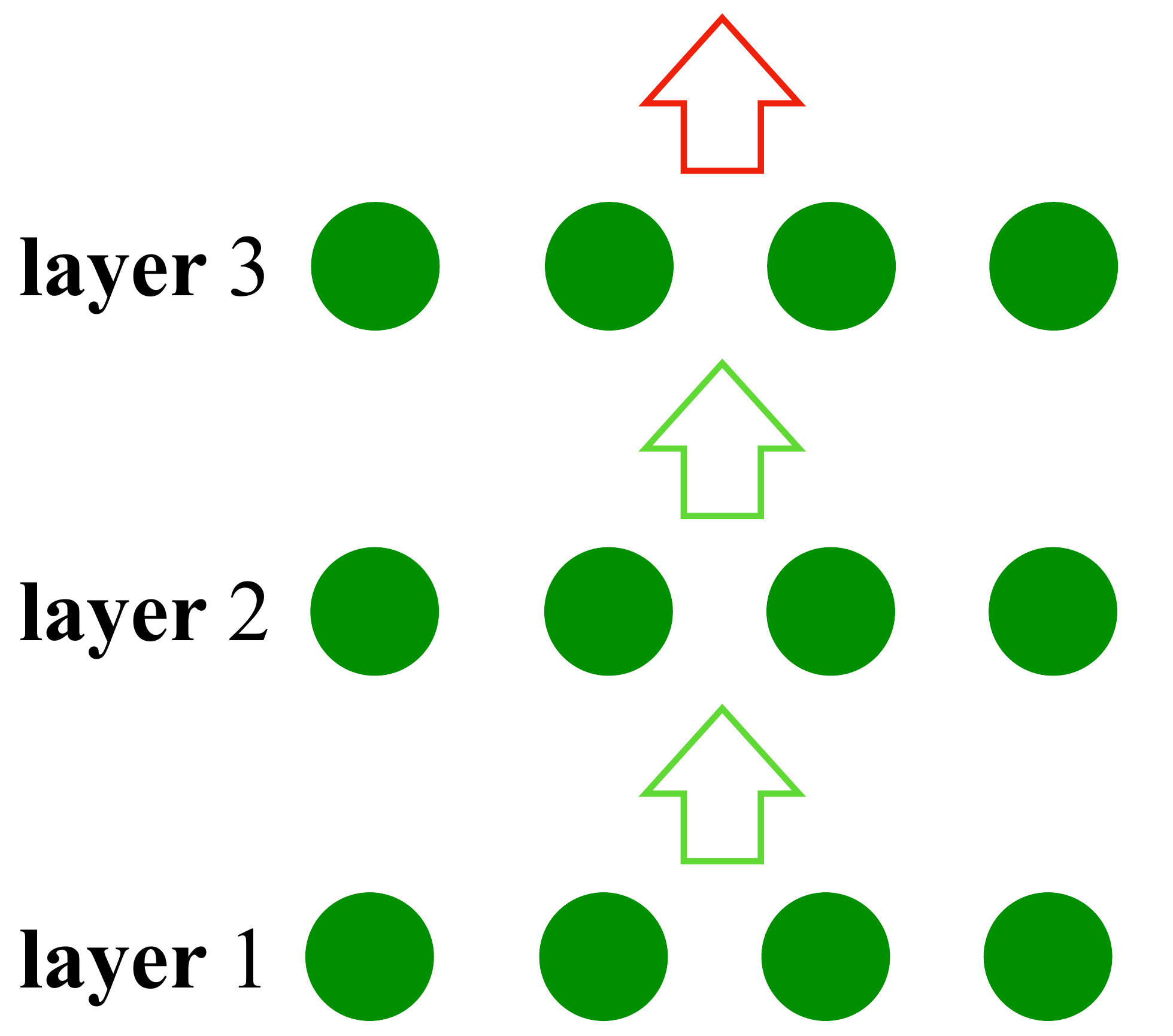}
\label{fig:vanilla}
} \hfill
\subfloat[{\em Shallow} Sentential Context]{
\includegraphics[height=0.20\textwidth]{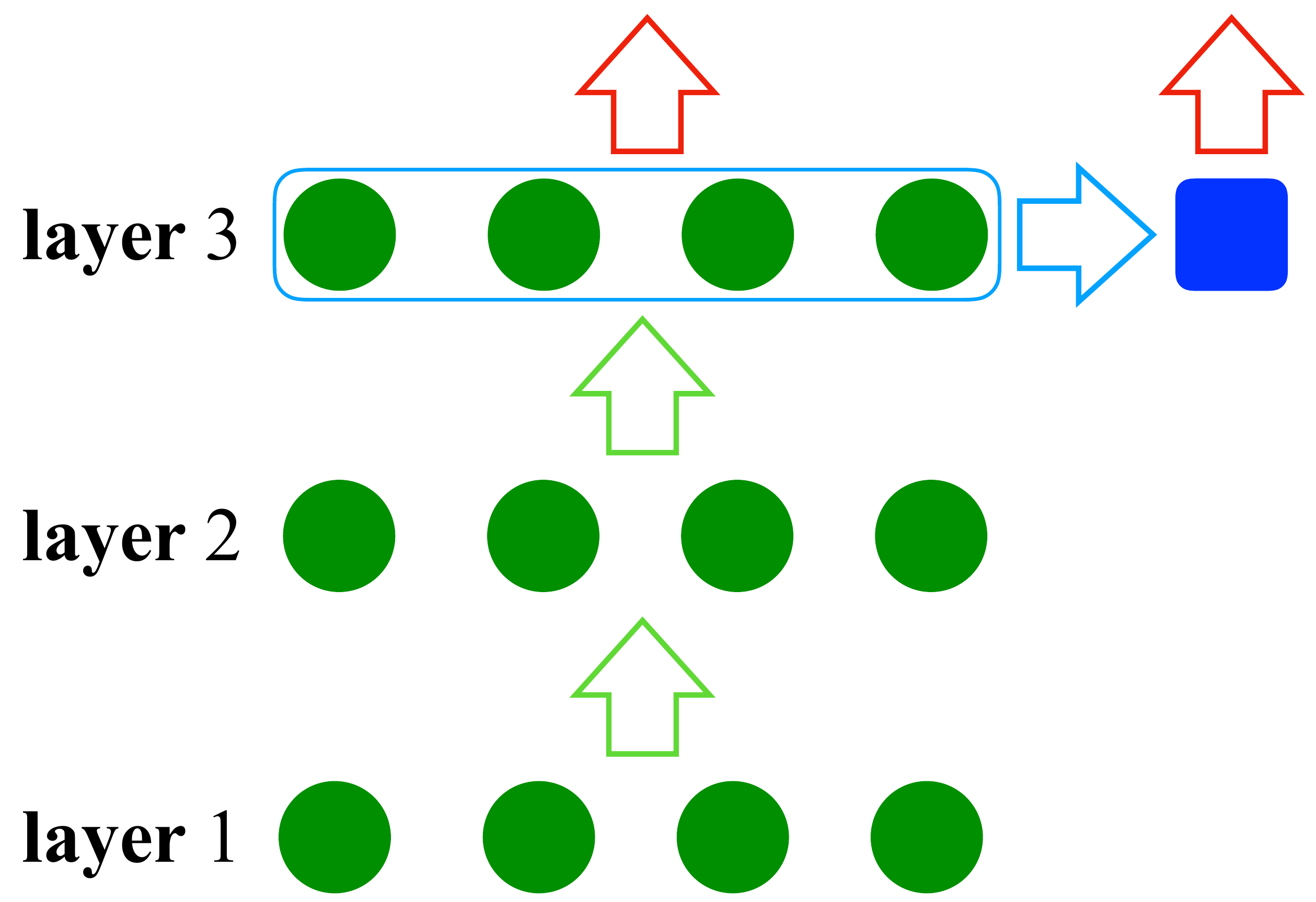}
\label{fig:global}
} \hfill
\subfloat[{\em Deep} Sentential Context ]{
\includegraphics[height=0.20\textwidth]{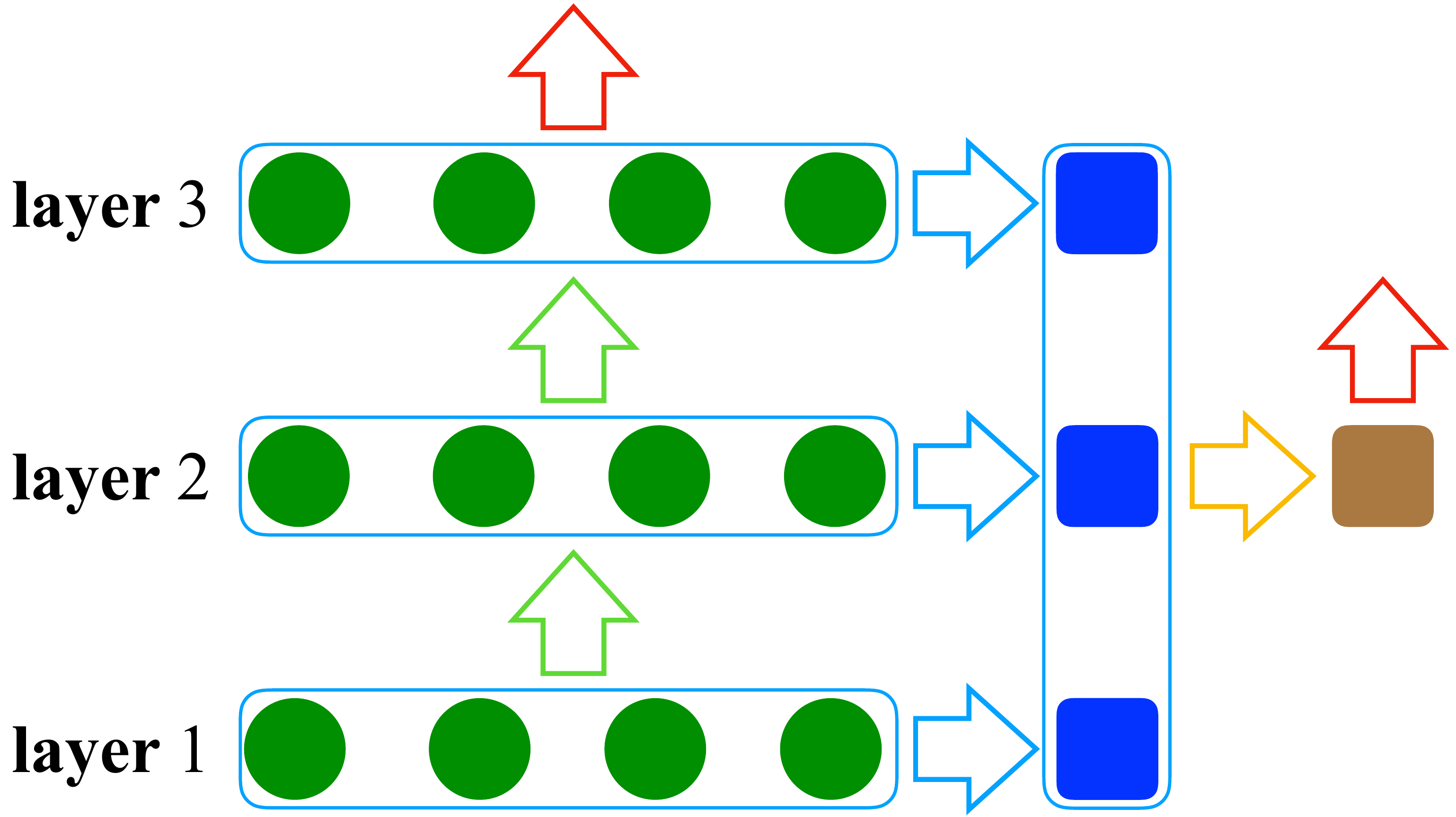}
\label{fig:global-deep}
}
\caption{Illustration of the proposed approache. As on a 3-layer encoder: (a) vanilla model without sentential context, (b) {\em shallow} sentential context representation (i.e. {\color{blue} blue square}) by exploiting the top encoder layer only; and (c) {\em deep} sentential context representation (i.e. {\color{brown} brown square}) by exploiting all encoder layers. The circles denote hidden states of individual tokens in the input sentence, and the squares denote the sentential context representations. The {\color{red} red} up arrows denote that the representations are fed to the subsequent decoder. This figure is best viewed in color.} 
\label{fig:proposed_models}
\end{figure*}

The contributions of this paper are:
\begin{itemize}
\item Our study demonstrates the necessity and effectiveness of exploiting source-side sentential context for NMT, which benefits from fusing useful contextual information across encoder layers.

\item We propose several strategies to better capture useful sentential context  for neural  machine  translation. Experimental  results   empirically   show   that   the   proposed approaches  achieve  improvement  over  the strong baseline model \textsc{Transformer}.
\end{itemize}

\section{Approach}

Like a human translator, the encoding process is analogous to reading a sentence in the source language and summarizing its meaning (i.e. sentential context) for generating the equivalents in the target language. When humans translate a source sentence, they generally scan the sentence to create a whole understanding, with which in mind they incrementally generate the target sentence by selecting parts of the source sentence to translate at each decoding step. 
In current NMT models, the attention model plays the role of selecting parts of the source sentence, but lacking a mechanism to guarantee that the decoder is aware of the whole meaning of the sentence. 
In response to this problem, we propose to augment NMT models with sentential context, which represents the whole meaning of the source sentence. 

\subsection{Framework}
Figure~\ref{fig:proposed_models} illustrates the framework of the proposed approach. Let ${\bf g} = g({\bf X})$ be the sentential context vector, and $g(\cdot)$ denotes the function to summarize the source sentence ${\bf X}$, which we will discuss in the next sections. 
There are many possible ways to integrate the sentential context into the decoder. The target of this paper is not to explore this whole space but simply to show that one fairly straightforward implementation works well and that sentential context helps.
In this work, we incorporate the sentential context into decoder as
\begin{eqnarray}
    {\bf d}^l_i &=& f(\textsc{Layer}_{dec}(\widehat{\bf D}^{l-1}), {\bf c}^l_i), \\
    \widehat{\bf D}^{l-1} &=& \textsc{Ffn}_l({\bf D}^{l-1}, {\bf g}),
\end{eqnarray}
where ${\bf d}^l_i$ is the $l$-th layer decoder state at decoding step $i$, ${\bf c}^l_i$ is a dynamic vector that selects certain parts of the encoder output, $\textsc{Ffn}_l(\cdot)$ is a distinct feed-forward network associated with the $l$-th layer of the decoder, which reads the $l-1$-th layer output ${\bf D}^{l-1}$ and the sentential context $\bf g$. In this way, at each decoding step $i$, the decoder is aware of the sentential context $\bf g$ embedded in $\widehat{\bf D}^{l-1}$. 

In the following sections, we discuss the choice of $g(\cdot)$, namely {\em shallow sentential context} (Figure~\ref{fig:global}) and {\em deep sentential context} (Figure~\ref{fig:global-deep}), which differ at the encoder layers to be exploited. It should be pointed out that the new parameters introduced in the proposed approach are jointly updated with NMT model parameters in an end-to-end manner.

\subsection{Shallow Sentential Context}

Shallow sentential context is a function of the top encoder layer output ${\bf H}^L$:
\begin{equation}
    {\bf g} = g({\bf H}^L) = \textsc{Global}({\bf H}^L),  \label{eqn:global}
\end{equation}
where $\textsc{Global}(\cdot)$ is the composition function.

\paragraph{Choices of \textsc{Global}($\cdot$)} Two intuitive choices are mean pooling~\cite{P15-1162} and max pooling~\cite{P14-1062}:
\begin{eqnarray}
    \textsc{Global}_{\textsc{mean}} &=& \textsc{Mean} ({\bf H}^L), \\
    \textsc{Global}_{\textsc{max}} &=& \textsc{Max} ({\bf H}^L).
\end{eqnarray}
Recently,~\newcite{lin2017structured} proposed a self-attention mechanism to form sentence representation, which is appealing for its flexibility on extracting implicit global features. Inspired by this, we propose an attentive mechanism to learn sentence representation:
\begin{eqnarray}
    \textsc{Global}_{\textsc{att}} &=& \textsc{Att}({\bf g}^0, {\bf H}^L), \\
    {\bf g}^0 &=& \textsc{Max}({\bf H}^0),
\end{eqnarray}
where ${\bf H}^0$ is the word embedding layer, and its max pooling vector ${\bf g}^0$ serves as the query to extract features to form the final sentential context representation.

\subsection{Deep Sentential Context}
Deep sentential context is a function of all encoder layers outputs $\{{\bf H}^1, \dots, {\bf H}^L\}$:
\begin{eqnarray}
    {\bf g} = g({\bf H}^1, \dots, {\bf H}^L) = \textsc{Deep}({\bf g}^1, \dots, {\bf g}^L),
\end{eqnarray}
where ${\bf g}^l$ is the sentence representation of the $l$-th layer ${\bf H}^l$, which is calculated by Equation~\ref{eqn:global}.
The motivation for this mechanism is that recent studies reveal that different encoder layers capture linguistic properties of the input sentence at different levels~\cite{Peters:2018:NAACL}, and aggregating layers to better fuse semantic information has proven to be of profound value~\cite{shen2018dense,Dou:2018:EMNLP,wang:2018:COLING,dou2019dynamic}. In this work, we propose to fuse the global information across layers.

\begin{figure}[t]
\centering
\subfloat[\textsc{Rnn}]{
\includegraphics[height=0.20\textwidth]{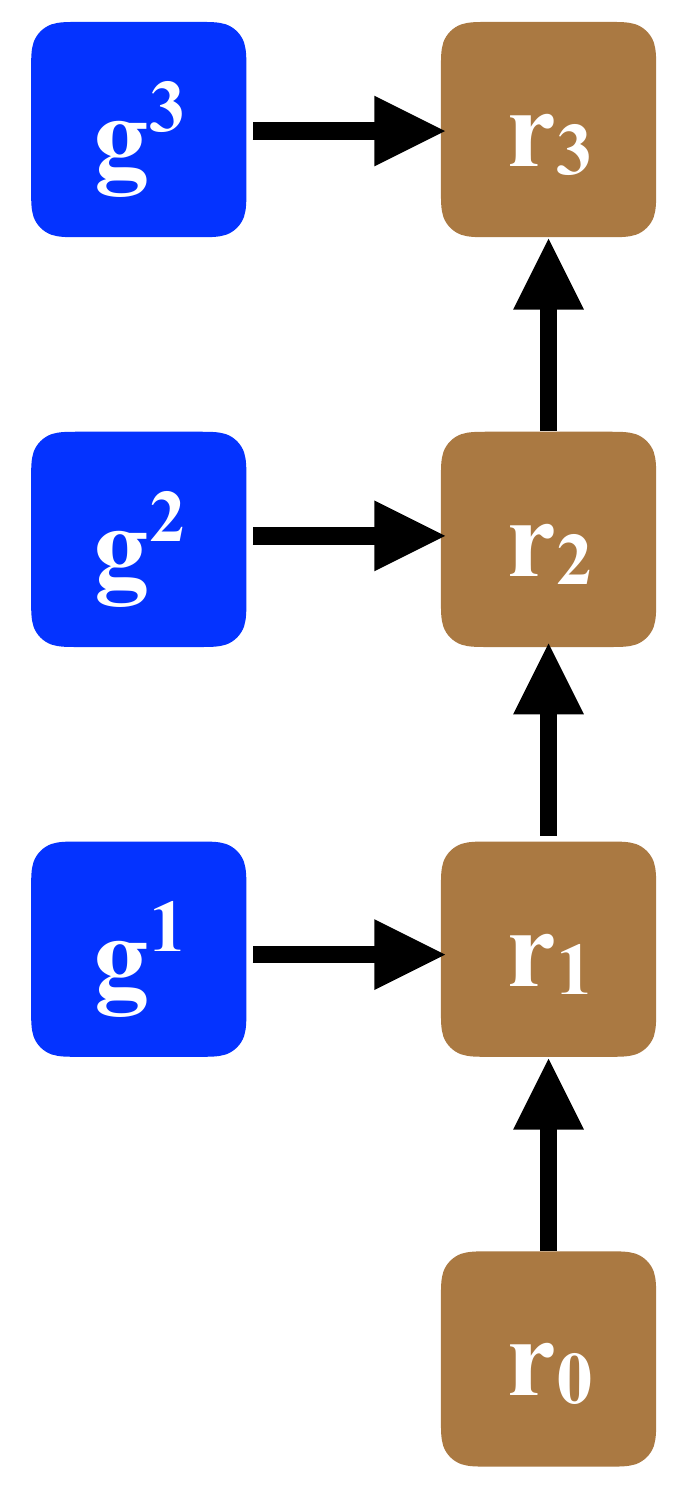}
\label{fig:rnn}
} \hspace{0.1\textwidth}
\subfloat[\textsc{Tam}]{
\includegraphics[height=0.20\textwidth]{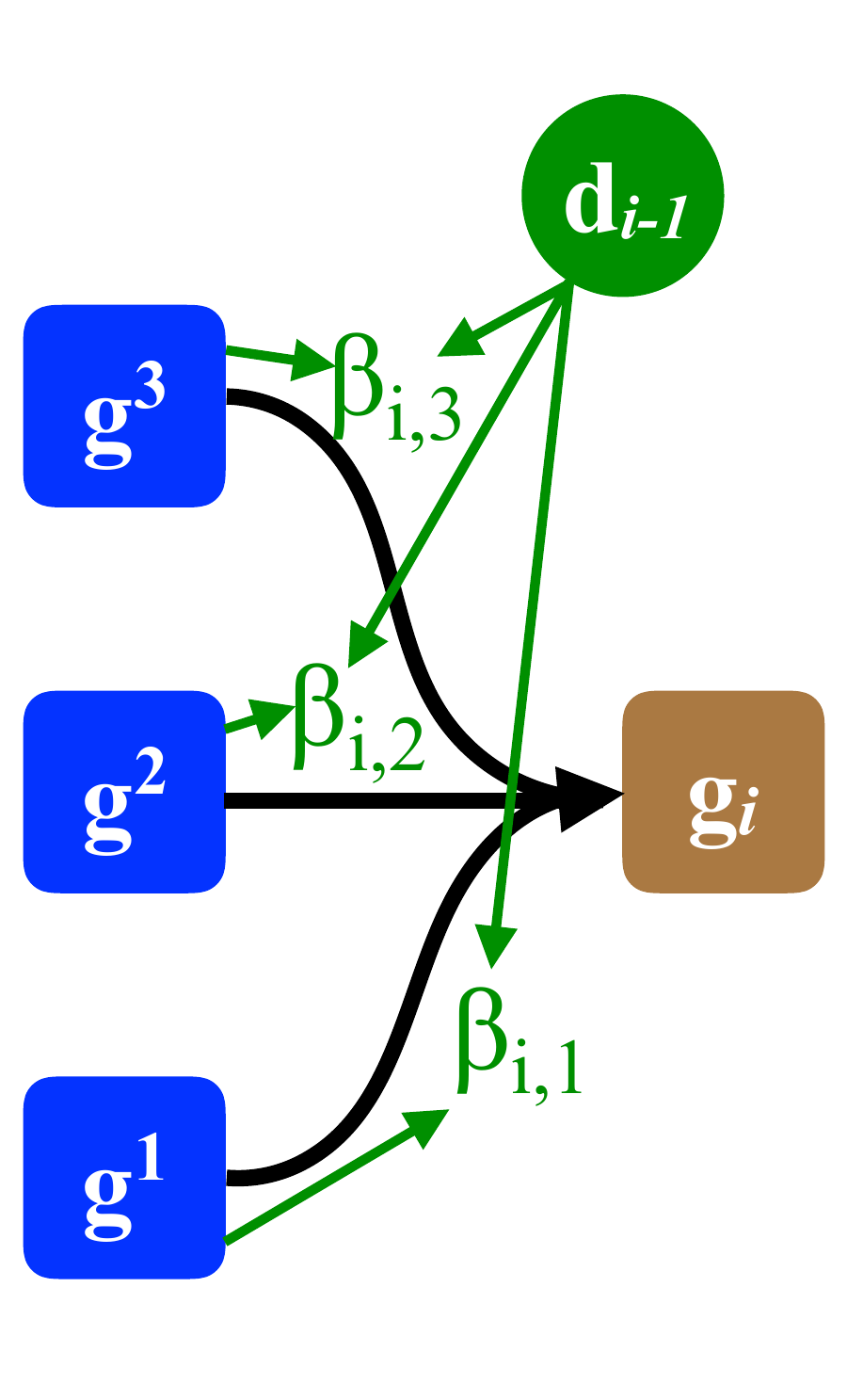}
\label{fig:tam}
}
\caption{Illustration of the deep functions. ``\textsc{Tam}'' model dynamically aggregates sentence representations at each decoding step with state ${\bf d}_{i-1}$.}
\label{fig:deep}
\end{figure}

\paragraph{Choices of \textsc{Deep}($\cdot$)}
In this work, we investigate two representative functions to aggregate information across layers, which differ at whether the decoding information is taken into account.

\noindent{\bf \em RNN} 
Intuitively, we can treat ${\bf G} = \{{\bf g}^1, \dots, {\bf g}^L\}$ as a sequence of representations, and recurring all the representations with an RNN:
\begin{eqnarray}
    \textsc{Deep}_{\textsc{Rnn}} = \textsc{Rnn}({\bf G}).    
\end{eqnarray}
We use the last RNN state as the sentence representation: ${\bf g} = {\bf r}_L$.
As seen, the RNN-based aggregation repeatedly revises the sentence representations of the sequence with each recurrent step. As a side effect coming together with the proposed approach, the added recurrent inductive bias of RNNs has proven beneficial for many sequence-to-sequence learning tasks such as machine translation~\cite{dehghani2018universal}.

\noindent{\bf \em TAM} Recently,~\newcite{D18-1338} proposed a novel {\em transparent attention model} (\textsc{Tam}) to train very deep NMT models. In this work, we apply \textsc{Tam} to aggregate sentence representations:
\begin{eqnarray}
    \textsc{Deep}_{\textsc{Tam}} &=& \sum_{l=1}^L \beta_{i,l} {\bf g}^l, \\
    {\bm \beta}_i &=& \textsc{Att}_g ({\bf d}^l_{i-1}, {\bf G}),
\end{eqnarray}
where $\textsc{Att}_g(\cdot)$ is an attention model with its own parameters, that specifics which context representations is relevant for each decoding output. Again, ${\bf d}^l_{i-1}$ is the decoder state in the $l$-th layer.

Comparing with its \textsc{Rnn} counterpart, the \textsc{Tam} mechanism has three appealing strengths. First, \textsc{Tam} dynamically generates the weights $\bm \beta_i$ based on the decoding information at every decoding step ${\bf d}^l_{i-1}$, while \textsc{Rnn} is unaware of the decoder states and the associated parameters are fixed after training.
Second, \textsc{Tam} allows the model to adjust the gradient flow to different layers in the encoder depending on its training phase.

\section{Experiment}

\begin{table*}[h]
  \centering
  \begin{tabular}{c|c|c|c||r|c|c||c}
    \bf \#   &   \bf Model     &     $\textsc{Global}(\cdot)$     &  $\bf \textsc{Deep}(\cdot)$    &  \bf {\# Para.} & \bf {Train}  & \bf{Decode} &   {BLEU}\\  
    \hline
    1   &   \textsc{Base}     &   n/a    &       n/a &    88.0M	    &   1.39   &    3.85   &   27.31\\
    2   & \textsc{Medium}     &   n/a    &       n/a &    +25.2M    &   1.08   &    3.09   &   27.81\\
    \hline
    \hline
    3   &   \multirow{3}{*}{\textsc{Shallow}}       &   Mean Pooling     &    \multirow{3}{*}{n/a}                                                   &      +18.9M  & 1.35    & 3.45 & 27.58 \\
    4   &                        &   Max Pooling &  &      +18.9M  & 1.34    & 3.43 & 27.81$^\uparrow$ \\
    5  &                         &   Attention &   &  +19.9M & 1.22 & 3.23 & 28.04$^\Uparrow$ \\
    \hline
    6   &   \multirow{2}{*}{\textsc{Deep}}     &   \multirow{2}{*}{Attention}   &     \textsc{Rnn}  &  +26.8M  & 1.03 & 3.14 & 28.38$^\Uparrow$  \\
    7   &                                    &                                       &     \textsc{Tam}  &  +26.4M  & 1.07 & 3.03 & 28.33$^\Uparrow$ \\
    
\end{tabular}
  \caption{Impact of components on WMT14  En$\Rightarrow$De translation task. BLEU scores in the table are case sensitive. ``Train'' denotes the training speed (steps/second), and ``Decode'' denotes the decoding speed (sentences/second) on a Tesla P40. ``TAM'' denotes the transparent attention model to implement the function \textsc{Deep}($\cdot$). ``$\uparrow/\Uparrow$'': significant over \textsc{Transformer} counterpart ($p < 0.05/0.01$), tested by bootstrap resampling~\cite{Koehn2004Statistical}. }
    \label{tab:component}
\end{table*}

We conducted experiments on WMT14 En$\Rightarrow$De and En$\Rightarrow$Fr benchmarks, which contain 4.5M and 35.5M sentence pairs respectively.
We reported experimental results with case-sensitive 4-gram BLEU score. 
We used byte-pair encoding (BPE)~\cite{sennrich2015neural} with 32K merge operations 
to alleviate the out-of-vocabulary problem. 
We implemented the proposed approaches on top of \textsc{Transformer} model \cite{Vaswani:2017:NIPS}. We followed~\newcite{Vaswani:2017:NIPS} to set the model configurations, and reproduced their reported results. 
We tested both \emph{Base} and \emph{Big} models, which differ at the layer size (512 vs. 1024) and the number of attention heads (8 vs. 16).  

\subsection{Ablation Study}
We first investigated the effect of components in the proposed approaches, as listed in Table~\ref{tab:component}. 

\paragraph{Shallow Sentential Context} (Rows 3-5)
All the shallow strategies achieve improvement over the baseline \emph{Base} model, validating the importance of sentential context in NMT. Among them, attentive mechanism (Row 5) obtains the best performance in terms of BLEU score, while maintains the training and decoding speeds. Therefore, we used the attentive mechanism to implement the function \textsc{Global}($\cdot$) as the default setting in the following experiments.

\paragraph{Deep Sentential Context} (Rows 6-7) 
As seen, both \textsc{Rnn} and \textsc{Tam} consistently outperform their shallow counterparts, proving the effectiveness of deep sentential context.
Introducing deep context significantly improves translation performance by over 1.0 BLEU point, while only marginally decreases the training and decoding speeds.

\paragraph{Compared to Strong Base Model} (Row 2) 
As our model has more parameters than the \emph{Base} model, we build a new baseline model (\textsc{Medium} in Table~\ref{tab:component}) which has a similar model size as the proposed deep sentential context model. We change the filter size from 1024 to 3072 in the decoder's feed-forward network (Eq.2). As seen, the proposed deep sentential context models also outperform the \textsc{Medium} model over 0.5 BLEU point.

\begin{table}[t]
\begin{center}
\begin{tabular}{c||c|c}
   {\bf Model}  &   {\bf En$\Rightarrow$De}  &   {\bf En$\Rightarrow$Fr}\\
    \hline \hline
    \textsc{Transformer-Base}    &   27.31  &   39.32 \\
    ~~~ +  \textsc{Deep} (\textsc{Rnn})       &  \bf 28.38$^\Uparrow$ & 40.15$^\Uparrow$ \\ 
    ~~~ +  \textsc{Deep} (\textsc{Tam})       &  28.33$^\Uparrow$  &  \bf 40.27$^\Uparrow$ \\ 
    \hline
    \textsc{Transformer-Big} & 28.58 &  41.41   \\  
    ~~~ +  \textsc{Deep} (\textsc{Rnn}) & 29.04$^\uparrow$ &  41.87 \\ 
    ~~~ +  \textsc{Deep} (\textsc{Tam}) & \bf 29.19$^\Uparrow$ & \bf 42.04$^\Uparrow$ \\ 
  \end{tabular}
  \caption{Case-sensitive BLEU scores on WMT14 En$\Rightarrow$De and En$\Rightarrow$Fr test sets. ``$\uparrow/\Uparrow$'': significant over \textsc{Transformer} counterpart ($p < 0.05/0.01$), tested by bootstrap resampling.} 
  \label{tab:exist}
  \end{center}
\end{table}

\subsection{Main Result} 
Experimental results on both WMT14 En$\Rightarrow$De and En$\Rightarrow$Fr translation tasks are listed in Table \ref{tab:exist}.
As seen, exploiting deep sentential context representation consistently improves translation performance across language pairs and model architectures, demonstrating the necessity and effectiveness of modeling sentential context for NMT. Among them, \textsc{Transformer-Base} with deep sentential context achieves comparable performance with the vanilla \textsc{Transformer-Big}, with only less than half of the parameters (114.4M vs. 264.1M, not shown in the table). 
Furthermore,  \textsc{Deep} (\textsc{Tam}) consistently outperforms \textsc{Deep} (\textsc{RNN}) in the \textsc{Transformer-Big} configuration. One possible reason is that the big models benefit more from the improved gradient flow with the transparent attention~\cite{D18-1338}.

\begin{table*}[t]
  \centering
  \renewcommand{\arraystretch}{1.2} 
  \scalebox{0.8}{
  \begin{tabular}{l||c c c| c c c c| c c c c c c}
    \multirow{2}{*}{\bf Model}   &  \multicolumn{3}{c|}{\bf Surface} & \multicolumn{4}{c|}{\bf Syntactic} &     \multicolumn{6}{c}{\bf Semantic}\\
    \cline{2-14}
     &\bf{SeLen} & \bf {WC} & \bf{Avg} & \bf {TrDep} & \bf {ToCo} & \bf {BShif} & \bf{Avg} & \bf {Tense} & \bf {SubN} & \bf {ObjN} & \bf {SoMo} & \bf {CoIn} & \bf{Avg} \\
    \hline
    \hline
    \multirow{1}{*}{\textsc{L4 in Base}}  & 94.18 & 66.24  & 80.21 & 43.91 & 77.36 & 69.25 & 63.51 & 88.03 & 83.77 & 83.68 & 52.22 & 60.57 & 73.65 \\
    \multirow{1}{*}{\textsc{L5 in Base}}  & 93.40 & 63.95  & 78.68 & 44.36 & 78.26 & 71.36 & 64.66 & 88.84 & 84.05 & 84.56 & 52.58 &  61.56 & 74.32 \\
    \hline
    \hline
    \textsc{L6 in Base}  & \bf92.20 & 63.00  & 77.60 & 44.74 & \bf79.02 & 71.24 & 65.00 & \bf89.24 & 84.69 & 84.53 & 52.13 & 62.47 & 74.61\\
    \hline
    {~~ + \textsc{Ssr}} 
     & 92.09 & 62.54 & 77.32 & 44.94 & 78.39 & 71.31 & 64.88 & 89.17 & 85.79 & \bf85.21 & 53.14 & 63.32 & 75.33\\
    \hline
    {~~ + \textsc{Dsr}} 
     & 91.86 & \bf65.61 & 78.74 & \bf45.52 & 78.77 & \bf71.62 & 65.30 & 89.08 & \bf 85.89 & 84.91 & \bf53.40 & \bf63.33 & 75.32 \\
  \end{tabular}
  } 
  \caption{Performance on the linguistic probing tasks of evaluating linguistics embedded in the encoder outputs. ``\textsc{Base}'' denotes the representations from \textsc{Tranformer-Based} encoder. ``\textsc{Ssr}'' denotes shallow sentence representation. ``\textsc{Dsr}'' denotes deep sentence representation. ``\textsc{Avg}'' denotes the average accuracy of each category.} 
  \label{tab:probing}
\end{table*}

\subsection{Linguistic Analysis}
To gain linguistic insights into the global and deep sentence representation, we conducted probing tasks\footnote{\url{https://github.com/facebookresearch/SentEval/tree/master/data/probing}}~\cite{P18-1198} to evaluate linguistics knowledge embedded in the encoder output and the sentence representation in the variations of the \emph{Base} model that are trained on En$\Rightarrow$De translation task. 
The probing tasks are classification problems that focus on simple linguistic properties of sentences. The 10 probing tasks are categories into three groups: (1) Surface information.  (2) Syntactic information. (3) Semantic information. 
For each task, we trained the classifier on the train set, and validated the classifier on the validation set. 
We followed~\newcite{hao-etal-2019-modeling} and~\newcite{li-etal-2019-information} to set the model configurations. 
We also listed the results of lower layer representations ($L=4, 5$) in \textsc{Transformer-Base} to conduct better comparison. 

The accuracy results on the different test sets are shown in Table~\ref{tab:probing}.  
From the tale, we can see that 
\begin{itemize}
\item For different encoder layers in the baseline model (see ``L4 in \textsc{Base}'', ``L5 in \textsc{Base}'' and ``L6 in \textsc{Base}''), lower layers embed more about surface information 
while higher layers encode more semantics, which are consistent with previous findings in \cite{W18-5431}.

\item Integrating the shallow sentence representation (``+ \textsc{Ssr}'') obtains improvement over the baseline on semantic tasks (75.33 vs. 74.61), while fails to improve on the surface (77.32 vs. 77.60) and syntactic tasks (64.88 vs. 65.00). This may indicate that the shallow representations that exploits only the top encoder layer (``L6 in \textsc{Base}'') encodes more semantic information. 

\item Introducing deep sentence representation  (``+ \textsc{Dsr}'') brings more improvements. The reason is that our deep sentence representation is induced from the  sentence  representations of all the encoder layers, and lower layers that contain abound surface and syntactic information are exploited.
\end{itemize}

Along with the above translation experiments, we believe that the sentential context is necessary for NMT by enriching the source sentence representation. The {\em deep sentential context} which is induced from all encoder layers can improve translation performance by offering different types of syntax and semantic information.

\section{Related Work}
Sentential context has been successfully applied in SMT~\cite{meng-EtAl:2015:ACL-IJCNLP, zhang2015local}. In these works,  sentential context representation which is generated by the CNNs is exploited to guided the target sentence generation. In broad terms, sentential context can be viewed as a sentence abstraction from a specific aspect. From this point of view, domain information~\cite{foster2007mixture,hasler2014combining,wang-etal-2017-sentence} and topic information~\cite{P12-1079,xiong2015topic,C16-1170} can also be treated as the sentential context, the exploitation of which we leave for future work. 

In the context of NMT, several researchers leverage document-level context for NMT~\cite{Wang:2017:EMNLP,choi2017context,Tu:2018:TACL}, while we opt for sentential context.
In addition, contextual information are used to improve the encoder representations~\cite{Yang:2018:EMNLP,Yang:2019:AAAI,C18-1276}. Our approach is complementary to theirs by better exploiting the encoder representations for the subsequent decoder. 
Concerning guiding the NMT generation with source-side context,~\newcite{zheng2018modeling} split the source content into translated and untranslated parts, while we focus on exploiting global sentence-level context.

\section{Conclusion}
In this work, we propose to exploit sentential context for neural machine translation. Specifically, the {\em shallow} and the {\em deep} strategies exploit the top encoder layer and all the encoder layers, respectively.  Experimental results on WMT14 benchmarks show that exploiting sentential context improves performances over the state-of-the-art \textsc{Transformer} model.
Linguistic analyses reveal that the proposed approach indeed captures more linguistic information as expected.

\balance
\bibliography{acl2019}
\bibliographystyle{acl_natbib}

\end{document}